\documentclass{article}
\usepackage{ijcai17}

\usepackage{times}
\usepackage{epsfig}
\usepackage{graphicx}
\usepackage{amsmath}
\usepackage{amssymb}
\usepackage{amsfonts}
\usepackage{multirow}
\usepackage{algorithmic}
\usepackage{array}
\usepackage{mdwmath}
\usepackage{mdwtab}
\usepackage{eqparbox}
\usepackage{url}
\usepackage{color}
\usepackage{colortbl}
\usepackage{subcaption}
\usepackage{bm}

\newcommand{\fref}[1]{Fig.~\ref{#1}}

\newcommand{\etc}[1]{~\textit{etc.}}
\newcommand{\etal}[1]{~\textit{et~al.}}

\title{Automatic Generation of Grounded Visual Questions}

\author{
Shijie Zhang{\small $~^{\#1}$},
Lizhen Qu\thanks{Corresponding author.}{\small $~^{\S\P2}$},
Shaodi You{\small $~^{\S\P3}$},
Zhenglu Yang{\small $~^{\dag 4}$},
Jiawan Zhang{\small $~^{\ddag 5}$} \\
$~^{\#}$School of Computer Science and Technology, Tianjin University, Tianjin, China  \\
$~^{\S}$Data61-CSIRO, Canberra, Australia  \\
$~^{\P}$Australian National University, Canberra, Australia  \\
$~^{\dag}$College of Computer and Control Engineering, College of Software, Nankai University, Tianjin, China \\
$~^{\ddag}$The School of Computer Software, Tianjin University, Tianjin, China  \\
$~^{1}~^{5}$\{shijiezhang,jwzhang\}@tju.edu.cn
$~^{2}$lizhen.qu@data61.csiro.au
$~^{3}$shaodi.you@anu.edu.au
$~^{4}$yangzl@nankai.edu.cn
}

\begin{document}
\maketitle

\begin{abstract}
In this paper, we propose the first model to be able to generate visually grounded questions with \textit{diverse} types for a single image. Visual question generation is an emerging topic which aims to ask questions in natural language based on visual input. To the best of our knowledge, it lacks automatic methods to generate meaningful questions with various types for the same visual input.

To circumvent the problem, we propose a model that automatically generates visually grounded questions with \textit{varying} types. Our model takes as input both images and the captions generated by a dense caption model, samples the most probable question types, and generates the questions in sequel. The experimental results on two real world datasets show that our model outperforms the strongest baseline in terms of both correctness and diversity with a wide margin. 
\end{abstract}


\section{Introduction}
Multi-modal learning of vision and language is an important task in artificial intelligence because it is the basis of many applications such as education, user query prediction, interactive navigation, and so forth. Apart from describing visual scenes by using declarative sentences~\cite{Chen14,Gupta12,Karpathy15,Hodosh13,kulkarni11,kuznetsova12,li09,vinyals15,xu2015show}, recently, automatic answering of visually related questions (VQA) has also attracted a lot of attention in computer vision communities~\cite{Antol15,malinowski14,gao2015you,ren15,yu15,zhu2015visual7w}. However, there is little work on automatic generation of questions for images.

\textit{"The art of proposing a question must be held higher value than solving it. -Georg Canton"}. An intelligent system should be able to ask meaningful questions given the environment. Beyond demonstrating a high-level of AI, in practice, multi-modal question-asking modules find their use in a wide range of AI systems such as child education and dialogue systems.

To the best of our knowledge, almost all existing VQA systems rely on manually constructed questions \cite{Antol15,malinowski14,gao2015you,ren15,yu15,zhu2015visual7w}. An common assumption of the existing VQA systems is that answers are visually grounded thus all relevant information can be found in the visual input. However, the construction of such data sets are labor-intensive and time consuming, thus limits the diversity and coverage of questions being asked. As a consequence, the data incompleteness imposes a special challenge for supervised-learning based VQA systems.


In light of the above analysis, we focus on automatic generation of visually grounded questions, coined VQG. The generated questions should be grammatically well-formed, reasonable for given images, and as diverse as possible. However, the existing systems are either rule-based such that they generate questions with few limited textual patterns \cite{ren15,zhu2015visual7w}, or they are able to ask only one question per image and the generated questions are frequently not visually grounded~\cite{MostafazadehMDZ16}.


\begin{figure}[!t]
\centering
\includegraphics[width=1\linewidth]{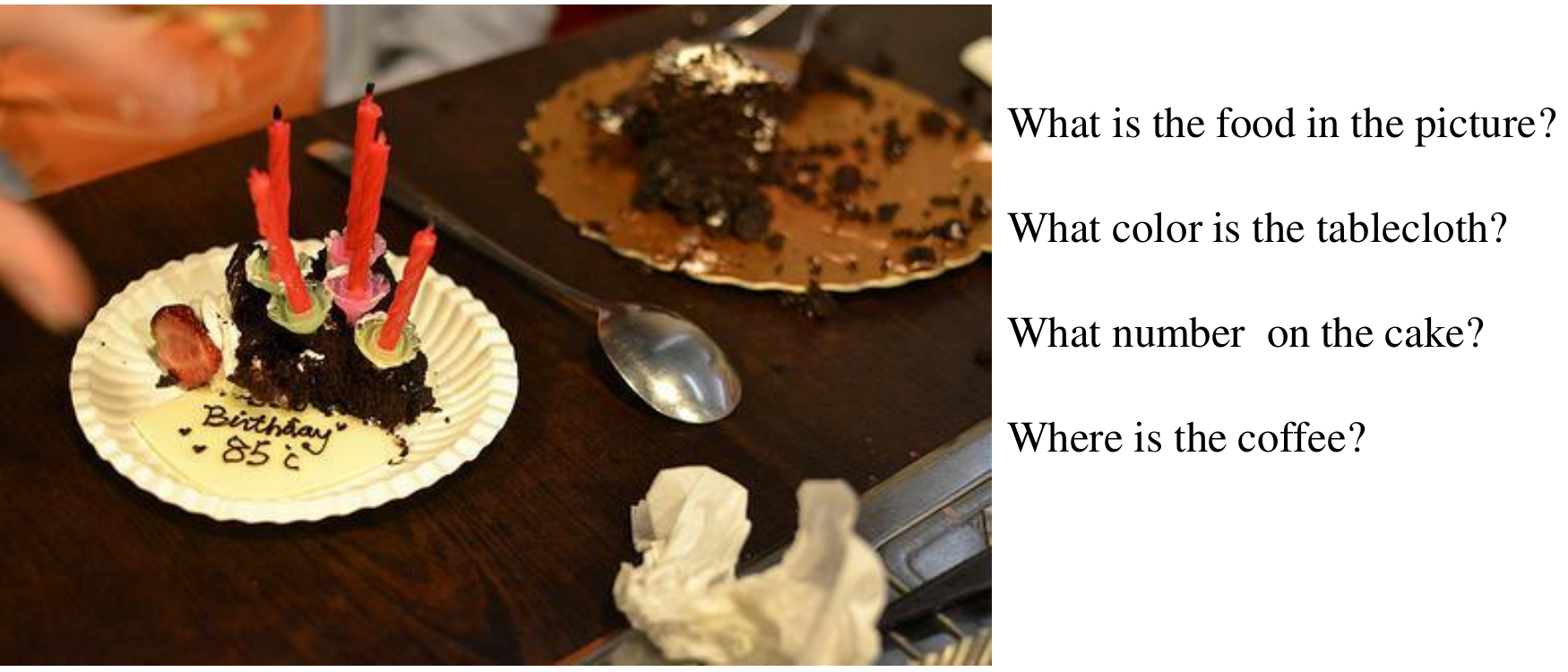}
\caption{Automatically generated grounded visual questions.}
\label{Fig:Teaser}
\end{figure}

\begin{figure*}[tbp]
\includegraphics[width=\linewidth]{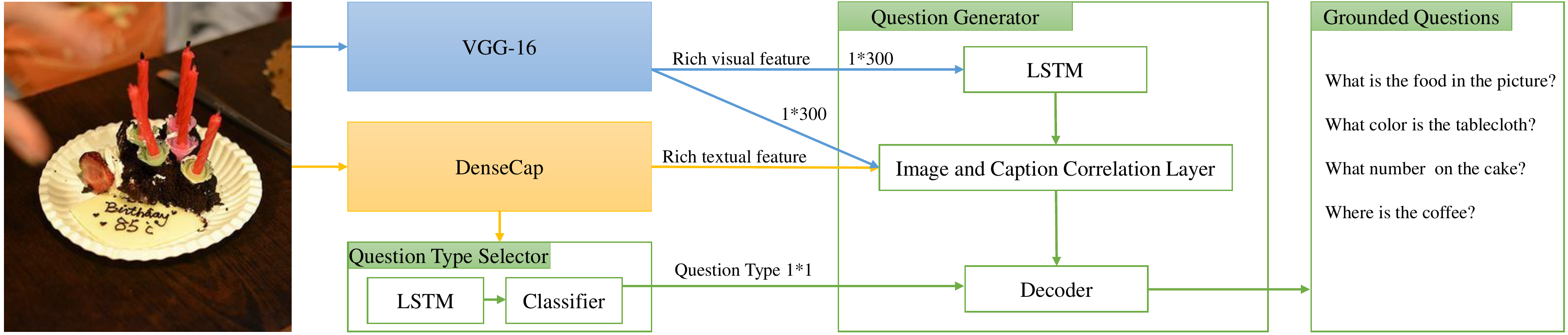}
\caption{The proposed framework.}
\label{Fig:Framework}
\vspace{-6pt}
\end{figure*}

To tackle this task, we propose the first model capable of asking questions of various types for the same image. As illustrated in \fref{Fig:Framework}, we first apply DenseCap \cite{johnson2015densecap} to construct dense captions that provides a almost complete coverage of information for questions. Then we feed these captions into the question type selector to sample the most probable question types. Taking as input the questions types, the dense captions, as well as visual features generated by VGG-16~\cite{Simonyan14c}, the question generator decodes all these information into questions. We conduct extensive experiments to evaluate our model as well as the most competitive baseline with three kinds of measures adapted from the ones commonly used in the tasks of image caption generation and machine translation.

The contributions of our paper are three-fold:

\begin{itemize}

\item We propose the first model capable of asking visually grounded questions with diverse types for a single image. 

\item Our model outperforms the strongest baseline up to 216\% in terms of the coverage of asked questions.

\item The grammaticality of the questions generated by our model as well as their relatedness to visual input also outperform the strongest baseline with a wide margin.

\end{itemize}

The rest of the paper is organized as follows: we cover the related work in Section \ref{sec:related}, followed by presenting our model in Section 3. After introducing the experimental setup in Section 4, we discuss the results in Section 5, and draw the conclusion in Section 6.

\section{Related Work}
\label{sec:related}

The generation of textual description for visual information has gained popularity in recent years. This includes joint learning of both visual information and text \cite{Barnard03,kong14,zitnick2013learning}.
A typical task is to describe images with a few declarative sentences, often referred to as image captions~\cite{Barnard03,Chen14,Gupta12,Karpathy15,Hodosh13,kulkarni11,kuznetsova12,li09,vinyals15,xu2015show}.

\vspace{-6pt}
\paragraph{Visual Question and Answering}
Automatic answering of questions based on visual input is an one of the most popular tasks in computer vision~\cite{Geman15,malinowski14,malinowski15,pirsiavash14,ren15,weston15,yu15}. VQA models are now been evaluated on a few datasets~\cite{Antol15,malinowski14,gao2015you,ren15,yu15,zhu2015visual7w}. For these datasets, while images are collected by sub-sampling MS-Coco \cite{lin14}, the questions-answer pairs are manually generated \cite{Antol15,gao2015you,yu15,zhu2015visual7w} or by using NLP tools\cite{ren15} that converts limited types of image captions into queries. 

\vspace{-6pt}
\paragraph{Visual Question Generation}
While asking questions automatically is explored in-depth in NLP, it is rarely researched for visually related questions. 
Such questions are strongly desired by creating VQA dataset.
Early methods simply converting image labeling into questions, which only allows generation of low-level questions. To diversify questions per image, however, it is still labor-consuming~\cite{Antol15,gao2015you,malinowski14}.
Zhu \etal \space \cite{zhu2015visual7w}, recently categorizes the manually generated questions into 7W question types, say, what, where, when and \etc.
Yu \etal \space \cite{yu15} consider question question as a task of selectively removing content words related answers from a caption. In a similar manner, Ren \etal \space \cite{ren15} design rules to transform image captions into questions with limited types.
Apart from that, the most closed work is~\cite{MostafazadehMDZ16} exploit abstract human like questions according to visual input. However, what they generate are ambiguous open questions where no determined answer is available within the visual input.
In a word, automatically generation of reasonable, and in the meanwhile, versatile close-form questions remains a challenging problem. 

\vspace{-6pt}
\paragraph{Knowledge Base (KB) based Question Answering (KB-QA)}
KB-QA has attracted considerable attention due to the ubiquity of the World Wide Web and the rapid development of the artificial intelligence (AI) technology. Large-scale structured KBs, such as DBpedia \cite{Auer2007DBpediaAN}, Freebase \cite{Bollacker2008FreebaseAC}, and YAGO \cite{Suchanek2007YagoAC}, provide abundant resources and rich general human knowledge, which can be used to respond to users' queries in open-domain question answering (QA). However, how to bridge the gap between visual questions and structured data in KBs remains a huge challenge. 

The existing KB-QA methods can be broadly classified into two main categories, namely, semantic parsing based methods\cite{Kwiatkowski2013,Reddy2016} and information retrieval based methods \cite{Yao2014InformationEO,bordes2014open} methods. Most semantic parsing based methods transform a question into its meaning representation (i.e., logical form), which will be then translated to a KB query to retrieve the correct answer(s). 
Information retrieval based methods initially roughly retrieve a set of candidate answers, and subsequently perform an in-depth analysis to re-rank the candidate answers and select the correct ones. These methods focus on modeling the correlation of question-answer pairs from the perspective of question topic, relation mapping, answer type, and so forth. 



\section{Question Generation}
\label{sec:qa_overall}
Our goal is to generate visually grounded questions directly from images with diverse question types. We start with randomly picking a caption from a set of automatically generated captions, which describes a certain region of image with natural language. Then we sample a reasonable question type and varying the caption. In the last step, our question generator learns the correlation between the caption and the image, generates a question of the chosen type. 

Formally, for each raw image $\mathbf{x}$, our model generates a set of captions $\{\mathbf{c}_1, \mathbf{c}_2, ..., \mathbf{c}_M \}$, samples a set of question types $\{\mathbf{t}_1, \mathbf{t}_2, ..., \mathbf{t}_{\hat{M}} \}$, followed by yielding a set of grounded questions $\{\mathbf{q}_1, \mathbf{q}_2, ..., \mathbf{q}_{\hat{M}} \}$. Herein, a caption or a question is a sequence of words.
\begin{equation}
\mathbf{w} = \{\mathbf{w}_1,...,\mathbf{w}_L \}
\end{equation}
Where $L$ is the length of the word sequence. Each word $\mathbf{w}_i$ employs 1-of-$K$ encoding, where $K$ is the size of the vocabulary. A question type is represented by the first word of a question, adopting 1-of-$T$ encoding where $T$ is the number of question types. The same as~\cite{zhu2015visual7w}, we consider six question types in our experiments: \textit{what}, \textit{when}, \textit{where}, \textit{who}, \textit{why} and \textit{how}.

For each image $\mathbf{x}_i$, we apply a dense caption model ({\textit{DenseCap})~\cite{johnson2015densecap} trained on the Visual Genome dataset~\cite{krishnavisualgenome} to produce a set of captions $\mathcal{C}_i$. Then the generative process is described as follows:
\begin{enumerate}
\item Choose a caption $\mathbf{c}_n$ from $\mathcal{C}_i$.
\item Choose a question type $\mathbf{t}_n$ given $\mathbf{c}_n$.
\item Generate a question $\mathbf{q}_n$ conditioned on $\mathbf{c}_n$ and $\mathbf{t}_n$.
\end{enumerate}

Denoted by $\bm{\theta}$ all model parameters, for each image $\mathbf{x}_i$, the joint distribution of $\mathbf{c}_n$, $\mathbf{t}_n$ and $\mathbf{q}_n$ is factorized as follows:
\begin{small}
\begin{align}\nonumber
P(\mathbf{q}_n,\mathbf{t}_n, \mathbf{c}_n| \mathbf{x}_i, \mathcal{C}_i; \bm{\theta}) =& P(\mathbf{q}_n | \mathbf{c}_n, \mathbf{x}_i, \mathbf{t}_n ; \bm{\theta}_q)\\
&P(\mathbf{t}_n |\mathbf{c}_n ; \bm{\theta}_t) P(\mathbf{c}_n|\mathcal{C}_i)
\end{align}
\end{small}
where $\bm{\theta} = \bm{\theta}_q \cup \bm{\theta}_t$, $P(\mathbf{q}_n | \mathbf{c}_n, \mathbf{x}_i, \mathbf{t}_n ; \bm{\theta}_q)$ is the distribution of generating question, $P(\mathbf{t}_n |\mathbf{c}_n ; \bm{\theta}_t)$ and $P(\mathbf{c}_n|\mathcal{C}_i)$ are the distributions for sampling question type and caption respectively. More details are given in the following sections. 

Since we do not observe the alignment between captions and questions, $\mathbf{c}_n$ is latent. Sum over $\mathbf{c}$, we obtain:
\begin{small}
\begin{displaymath}
P(\mathbf{q}_n, \mathbf{t}_n| \mathbf{x}_i, \mathcal{C}_i; \bm{\theta}) = \sum_{\mathbf{c}_n \in \mathcal{C}_i} P(\mathbf{q}_n,\mathbf{t}_n, \mathbf{c}_n| \mathbf{x}_i, \mathcal{C}_i; \bm{\theta})
\end{displaymath}
\end{small}
Let $\mathcal{Q}_i$ denote the question set of the image $\mathbf{x}_i$, the probability of the training dataset $\mathcal{D}$ is given by taking the product of the above probabilities over all images and their questions.
\begin{small}
\begin{equation}
\label{eq:likelihood}
P(\mathcal{D} | \bm{\theta}) = \prod_i \prod_{n \in \mathcal{Q}_i} P(\mathbf{q}_n, \mathbf{t}_n| \mathbf{x}_i, \mathcal{C}_i; \bm{\theta})
\end{equation}
\end{small}
For word representations, we initialize a word embedding matrix $\mathbf{E} \in \mathcal{R}^{300 \times K}$ by using Glove~\cite{pennington2014glove}, which are trained on 840 billions of words. For the image representations, we apply a VGG-16 model~\cite{szegedy2015going} trained on ImageNet~\cite{deng2009imagenet} without fine-tuning to produce 300-dimensional feature vectors. The dimension is chosen to match the size of the pre-trained word embeddings.

Compared to the question generation model~\cite{MostafazadehMDZ16}, which generates only one question per image, the probabilistic nature of this model allows generating questions of multiple types which refer to different regions of interests, because each caption predicted by \textit{DenseCap} is associated with a different region.

\subsection{Sample captions and question types}
The caption model \textit{DenseCap} generates a set of captions for a given image. Each caption $c$ is associated with a region and a confidence $o_c$ of the proposed region. Intuitively, we should give a higher probability to the caption with higher confidence than the lower one. Thus, given a caption set $\mathcal{C}_i$ of an image $\mathbf{x}_i$, we define the prior distribution as:
\begin{displaymath}
P(\mathbf{c}_k | \mathcal{C}_i) = \frac{\exp(o_k)}{\sum_{j \in \mathcal{C}_i} \exp(o_j) } 
\end{displaymath}

A caption is either a declarative sentence, a word, or a phrase. We are able to ask many different types of questions but not all of them for a chosen caption. For example, for a caption "floor is brown" we can ask "what color is the floor" but it would be awkward to ask a \emph{who} question. Thus, our model draws a question type given a caption with the probability $P(\mathbf{t}_n |\mathbf{c}_n)$ by assuming it suffices to infer question types given a caption. 

Our key idea is to learn the association between question types and key words/phrases in captions. The model $P(\mathbf{t}_n |\mathbf{c}_n)$ consists of two components. The first one is a Long Short Term Memory (LSTM)~\cite{hochreiter1997long} that maps a caption into a hidden representation. LSTM is a recurrent neural network taking the following form:
\begin{equation}
\label{eq:lstm}
\mathbf{h}_{t}, \mathbf{m}_{t}=\text{LSTM}(\mathbf{x}_t, \mathbf{h}_{t-1}, \mathbf{m}_{t-1})
\end{equation}
where $\mathbf{x}_t$ is the input and the hidden state of LSTM at time step $t$, and $\mathbf{h}_{t}$ and $\mathbf{m}_t$ are the hidden states and memory states of LSTM at time step $t$, respectively. As the representation of the whole sequence, we take the last state $\mathbf{h}_L$ generated at the end of the sequence. This representation is further fed into a softmax layer to compute a probability vector $\mathbf{p}_t$ for all question types. The probability vector characterizes a multinomial distribution of all question types.  

\subsection{Generate questions}
At the core of our model is the question generation module, which models $P(\mathbf{q}_n | \mathbf{c}_n, \mathbf{x}_i, \mathbf{t}_n; \bm{\theta}_q)$, given a chosen caption $\mathbf{c}_n$ and a question type $\mathbf{t}_n$. It is composed of three modules: i) an LSTM encoder to generate caption embeddings; ii) a correlation module to learn the association between images and captions; iii) a decoder consisting of an LSTM decoder and an ngram language model.

A grounded question is deeply anchored in both the sampled caption and the associated image. In our preliminary experiments, we found it useful to let the LSTM encoder $\text{LSTM}(\mathbf{x}_t, \mathbf{h}_{t-1}, \mathbf{m}_{t-1})$ to read the image features prior to reading captions. In particular, at time step $t = 0$, we initialize the state vector $\mathbf{m}_{0}$ to zero and feed the image features as $\mathbf{x}_0$. At the $1$st time step, the encoder reads in a special token $S_0$ indicating the start of a sentence, which is a good practice adopted by many caption generation models~\cite{vinyals15}. After reading the whole caption of length $L$, the encoder yields the last state vector $\mathbf{m}_L$ as the embedding of caption. 

The correlation module takes as input the caption embeddings from the encoder and the image features from VGG-16, produces a 300-dimensional joint feature map. We apply a linear layer of size $300 \times 600$ and a PReLU~\cite{he2015delving} layer in sequel to learn the associations between captions and images. Since an image gives an overall context and the chosen caption provides the focus in the image, the joint representation provides sufficient context to generate grounded questions. Although the LSTM encoder incorporates image features before reading captions, this correlation module enhances the correlation between images and text by building more abstract representations.

Our decoder extends the LSTM decoder of~\cite{vinyals15} with a ngram language model. The LSTM decoder consists of an LSTM layer and a softmax layer. The LSTM layer starts with reading the joint feature map and the start token $S_0$ in the same fashion as the caption encoder. From time step $t = 0$, the softmax layer predicts the most likely word given the
state vector at time $t$ yielded by the LSTM layer. A word sequence ends when the end of sequence token is produced. 

\paragraph{Joint decoding} Although the LSTM decoder alone can generate questions, we found that it would frequently produce repeated words and phrases such as "\emph{the the}". The problem didn't disappear even the beam search~\cite{Koehn2003} was applied. It is due to the fact that the state vectors produced at adjunct time steps tend to be similar. Since repeated words and phrases are rarely observed in text corpora, we discount such occurrence by joint decoding with a ngram language model. Given a word sequence $\mathbf{w} = \{w_1, ..., w_N \}$, a bigram language model is defined as:
\begin{displaymath}
P(\mathbf{w}) = \prod_{i = 2}^N P(w_{i} | w_{i - 1}) P(w_0)
\end{displaymath}
Instead of using neural models, we adopt the word count based estimation of model parameters. In particular, we apply the Kneser–Ney smoothing~\cite{kneser1995improved} to estimate $P(w_{i} | w_{i - 1})$, which is given by:
\begin{displaymath}
\frac{\max(\text{count}(w_{i - 1}, w_i) - d, 0)}{\text{count}(w_i)} + \lambda(w_{t-1})P_{KN}(w_i)
\end{displaymath}
where $\text{count}(x)$ denotes the corpus frequency of term $x$, $P_{KN}(w_i)$ is a back-off statistic of unigram $w_i$ in case the bigram $(w_i, w_{i-1})$ does not appear in the training corpus. The parameter $d$ is usually fixed to 0.75 to avoid overfitting for low frequency bigrams. And $\lambda(w_{t-1})$ is a normalizing constant conditioned on $w_i$. 

We incorporate bigram statistics with the LSTM decoder from the time step $t=1$ because the LSTM decoder can well predict the first words of questions. The LSTM decoder essentially captures the conditional probability $P_{l}(\mathbf{q}_t | \mathbf{q}_{< t})$, while the bigram model considers only the previous word $P_{b}(\mathbf{q}_t | \mathbf{q}_{t - 1})$ by using word counts. By interpolating these two, we obtain the final probability as:
\begin{displaymath}
P(\mathbf{q}_t | \mathbf{q}_{< t}) = (1 - \beta) P_l(\mathbf{q}_t | \mathbf{q}_{< t}) +\beta P_b(\mathbf{q}_t | \mathbf{q}_{t - 1}) 
\end{displaymath}
where $\beta \in [0 , 1]$ is an interpolation weight. In addition, we fix the first $k$ words of questions during decoding according to the chosen question types.

\subsection{Training}
\label{sec:train}
The key challenge of training is the involvement of the latent variables indicating the alignment between captions and gold standard questions for a deep neural network. We estimate the latent variables in a similar fashion as EM but computationally more efficient.

Suppose we are given the training set $\{(\mathbf{x}_1, \mathbf{q}_1), ...,  (\mathbf{x}_N, \mathbf{q}_N) \}$, the loss is given by:
\begin{displaymath}
l(\bm{\theta}) = \sum_{i=1}^N \sum_{n \in \mathcal{Q}_i} -\log P(\mathbf{q}_n, \mathbf{t}_n| \mathbf{x}_i, \mathcal{C}_i;  \bm{\theta})
\end{displaymath}
Suppose $Q(\mathbf{c}_n)$ denote some proposed distribution such that $\sum_n Q(\mathbf{c}_n) =1$ and $Q(\mathbf{c}_n) \geq 0$. Consider the following:

\begin{small}
\begin{align}\nonumber
\log P(\mathbf{q}_n, \mathbf{t}_n| \mathbf{x}_i, \mathcal{C}_i;  \bm{\theta})&=&\log \sum_{\mathbf{c}_k in \mathcal{C}_i} Q(\mathbf{c}_k)\frac{P(\mathbf{q}_n,\mathbf{t}_n, \mathbf{c}_k| \mathbf{x}_i, \mathcal{C}_i; \bm{\theta})}{Q(\mathbf{c}_k)}\\
&\geq& \sum_{\mathbf{c}_k in \mathcal{C}_i}Q(\mathbf{c}_k)\log \frac{P(\mathbf{q}_n,\mathbf{t}_n, \mathbf{c}_k| \mathbf{x}_i, \mathcal{C}_i; \bm{\theta})}{Q(\mathbf{c}_k)}
\label{eq:lower_bound}
\end{align}
\end{small}
The last step used Jensen’s inequality. The Equation \eqref{eq:lower_bound} gives a upper bound of the loss $l(\bm{\theta})$. When the bound is tight, we have $Q(\mathbf{c}_k) = P(\mathbf{c}_k | \mathbf{q}_n, \mathbf{t}_n ;  \bm{\theta})$. 

To save the EM loop, we propose a non-parametric estimation of $P(\mathbf{c}_k | \mathbf{q}_n, \mathbf{t}_n ;  \bm{\theta})$. As a result, for each question-image pair $(\mathbf{x}_n, \mathbf{q}_n)$, we maximize the lower bound by optimizing:
\begin{small}
\begin{displaymath}
\arg \min_{\bm{\theta}} - P(\mathbf{c}_n|\mathbf{q}_n, \mathbf{t}_n;  \bm{\theta}_c) \log [P(\mathbf{q}_n | \mathbf{c}_n, \mathbf{x}_n, \mathbf{t}_n ;  \bm{\theta}_q)P(\mathbf{t}_n | \mathbf{c}_n;\bm{\theta}_t)] + \text{const}
\label{eq:likelihood_em}
\end{displaymath}
\end{small}
This in fact assigns a weight $P(\mathbf{c}_n|\mathbf{q}_n, \mathbf{t}_n ;\bm{\theta})$ to each instance. By using a non-parametric estimation, we are still able to apply BackProp and the SGD style optimizing algorithms by just augmenting each instance with an estimated weight.  
 
Given a question $\mathbf{q}$ and a caption set $\mathcal{C}$ from the train set, we estimate $P(\mathbf{c}_k|\mathbf{q}, \mathbf{t};\bm{\theta})$ by using the kernel density estimator~\cite{scott2008kernel}:
\begin{equation}
P(\mathbf{c}_k|\mathbf{q}, \mathbf{t};\bm{\theta}) = P(\mathbf{c}_k|\mathbf{q};\bm{\theta}) = \frac{s(\mathbf{q}, \mathbf{c}_k)}{\sum_{\mathbf{c}_j \in \mathcal{C}}s(\mathbf{q}, \mathbf{c}_j)}
\label{eq:density}
\end{equation}
where $s(\mathbf{q}, \mathbf{c})$ is a similarity function between a question and a caption. We assume $\mathbf{c}_k$ are conditionally independent of $\mathbf{t}$ because we can directly extract the question type from the question $\mathbf{q}$ by looking at the first few words.

For a given question, there are usually very few matched captions generated by \textit{DenseCap} , hence the distribution of captions given a question is highly skewed. It is sufficient to randomly draw a caption each time to compute the probability based on Equation \eqref{eq:density}.

We formulate the similarity between a question and a caption by using both string similarity and embedding based similarity measures.

The surface string of a caption could be an exact or partial match of a given question. Thus we employ the Jaccard Index as string similarity measure between the surface string of a caption and that of a question.
\begin{displaymath}
\text{sim}_{s}(q, c) = \frac{q \cap c}{q \cup c} 
\end{displaymath}
where $c$ and $q$ denote their surface string respectively. Both strings are broken down to a set of char-based trigrams during the computation so that this measure still gives a high similarity if two strings differ only in some small variations such as singular and plural forms of nouns. 

In case of synonyms or words of similar meanings come with different form such as "car" and "automobile", we adopt the pre-trained word embeddings to calculate their similarity by using the weighted averaged of word embeddings:
\begin{small}
\begin{displaymath}
\text{sim}_{e}(\mathbf{q}, \mathbf{c}) = \text{cos}(\sum_{\mathbf{w}_i \in \mathbf{q}}\frac{\text{IDF}(\mathbf{w}_i)}{\sum_j \text{IDF}(\mathbf{w}_j)}\mathbf{E}\mathbf{w}_i, \sum_{\mathbf{w}_k \in \mathbf{c}}\frac{\text{IDF}(\mathbf{w}_k)}{\sum_j \text{IDF}(\mathbf{w}_j)}\mathbf{E}\mathbf{w}_k)
\end{displaymath}
\end{small}
where $\text{cos}$ denotes the cosine similarity, $\text{IDF}(x)$ is the inverse document frequency of word $x$ defined by $\frac{|\mathcal{V}|}{|\{d \in \mathcal{D}: x \in d\}|}$, and $\mathcal{D}$ is the corpus containing all questions, answers, and captions.

The final similarity measure is computed as the interpolation of the two measures:
\begin{displaymath}
s(\mathbf{q}, \mathbf{c}) = \alpha \text{sim}_s(\mathbf{q}, \mathbf{c}) + (1 - \alpha) \text{sim}_e(\mathbf{q}, \mathbf{c})
\end{displaymath}
where the hyperparameter $\alpha \in (0, 1)$.

\begin{figure*}[!t]
    \centering
    \includegraphics[width=\linewidth]{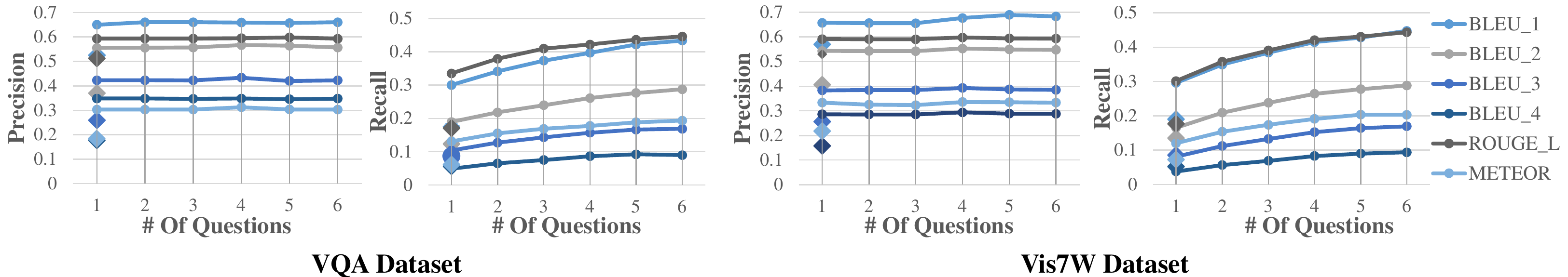}
    \vspace{-12pt}
    \caption{Precision and Recall of our method and baseline method(Neural Talk 2) on VQA Dataset and Visual7W Dataset. Results of our method are depicted with line with circles, we show the results by varying number of questions from one to six. By generating more questions, the Recall increase without a drop on precision. The baseline method (diamonds), however, can generate only one question per image.}
    \label{fig:PR_all}
    \vspace{-12pt}
\end{figure*}

\section{Experimental Setup}

\subsection{Datasets}
We conduct our experiments on two datasets: VQA-Dataset \cite{Antol15} and Visual7W \cite{zhu2015visual7w}. The former is the most popular benchmark for VQA and the latter is a recently created dataset with more visually grounded questions per image than VQA.

\noindent\textbf{VQA}: a sample from the MS-COCO dataset~\cite{lin14}, which contains 254,721 images and 764,163~manually compiled questions respectively. Each image is associated with three questions on average.

\noindent\textbf{Visual7W}: a dataset composed of 327,939 QA pairs on 47,300 COCO images, collected from the MS-COCO dataset~\cite{lin14} as well. In addition, it includes 1,311,756 human-generated answers in form of multiple-choice and 561,459 object groundings from 36,579 categories. 
Each image is associated with five questions on average.

\subsection{Baseline}
In this paper, we consider a baseline by training the image caption generation model NeuralTalk2~\cite{vinyals15} on image-question pairs. The baseline is almost the same as \cite{MostafazadehMDZ16}, which is the only work generating questions from visual input. The model of neuraltalk2 differs from \cite{MostafazadehMDZ16} only in the RNNs used in the decoder. NeuralTalk2 adopts LSTM while \cite{MostafazadehMDZ16} chooses GRU~\cite{cho2014learning}. The two RNN models achieve almost identical performance in language modeling~\cite{chung2015gated}.



\subsection{Evaluation measures}
As a common practice for evaluating generated word sequences we employ three different evaluation metrics: BLEU \cite{Papineni02bleu:a}, METEOR~\cite{banerjee2005meteor} and ROUGE-L~\cite{lin2004rouge}.

BLEU is a modified n-gram precision. We varied the size of ngram from one to four, computed the corresponding measures respectively for each image and averaged the results across all images. Both METEOR and ROUGE-L\footnote{We take the same $\beta$ of F-Measure as the implementation in https://github.com/tylin/coco-caption.} are F-Measures favoring precision, computed against the reference question with the highest score among all reference questions in the same image. The measures are averaged in sequel across all images. Therefore, all three of them are precision-oriented measures. 

To measure the diversity of our generated questions, we also compute the the same set of evaluation measures by comparing each reference sentence with the best matching generated sentence of the same images. This provides an estimate of coverage in analogy of recall.

\subsection{Implementation details}~
We optimize all models with Adam~\cite{kingma2014adam}. We fix the batch size to 64. We set the maximal epochs to 64 for Visual7W and the maximal epochs to 128 for VQA. The corresponding model hyperparameters were tuned on the validation sets. Herein, we set $\alpha = 0.75$.

\begin{figure*}[tbp]
\centering\includegraphics[width=1.0\linewidth]{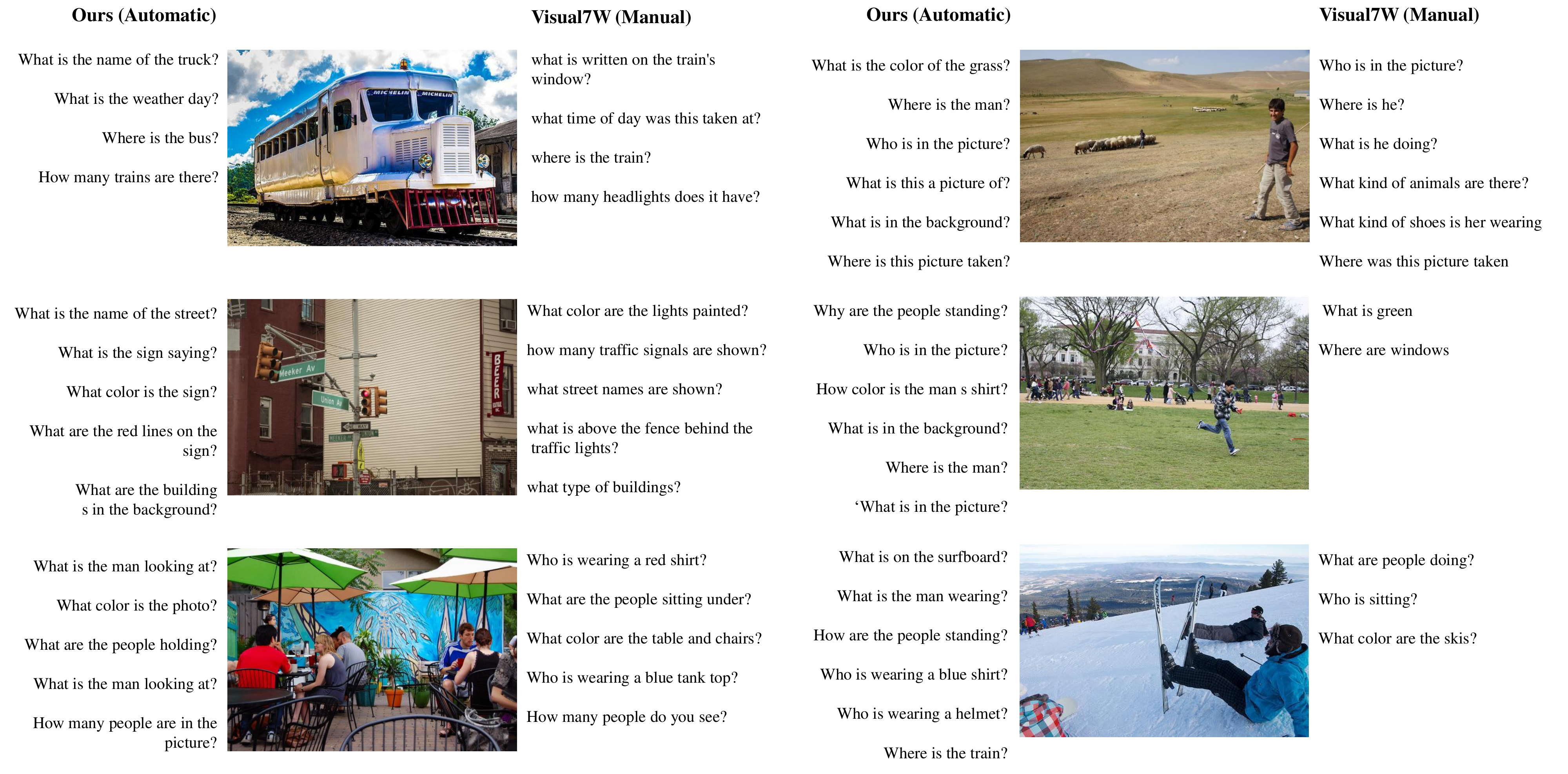}
\vspace{-12pt}
\caption{Comparison between manually composed questions and automatically generated questions by our method. Our method generates reasonable grounded questions and more versatile the manually provided ones.}
\label{Fig:Visual}
\vspace{-12pt}
\end{figure*}

\begin{figure*}[tbp]
\centering
\includegraphics[width=0.95\linewidth]{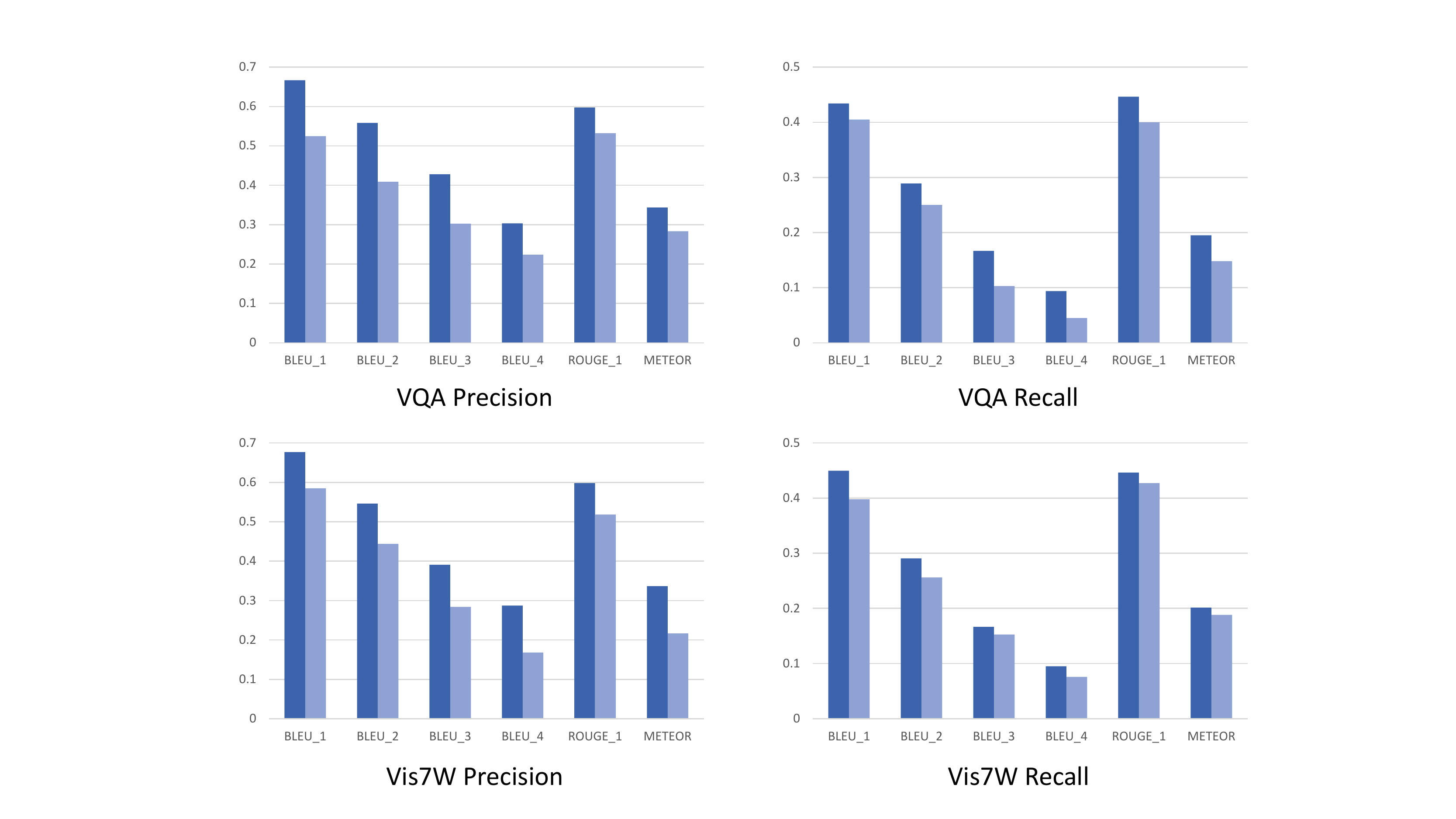}
\vspace{-12pt}
\caption{Comparing decoding between with the bigram model (dark blue) and without the bigram model (light blue).}
\label{Fig:Ngrams}
\vspace{-12pt}
\end{figure*}

\begin{figure*}[tb]
\centering
\includegraphics[width=0.75\linewidth]{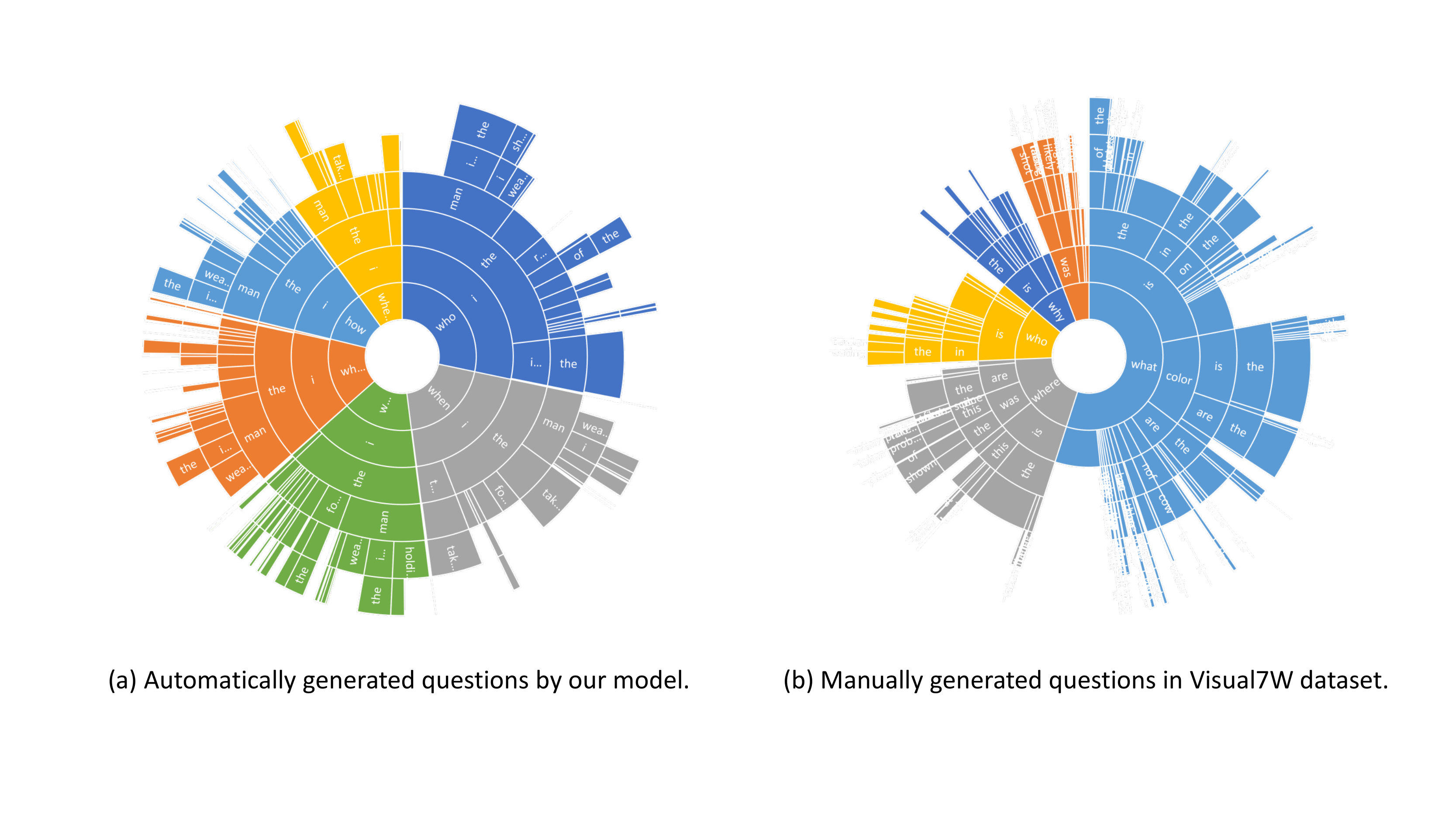}
\vspace{-12pt}
\caption{Distribution of question types.}
\label{Fig:question_types_dist}
\vspace{-12pt}
\end{figure*}

\section{Results and Discussions}
Figure \ref{fig:PR_all} illustrates all three precision-oriented measures evaluated on Visual7W and VQA datasets respectively. Our baseline is able to generate only one question per image. When we compare its results with the highest scored question per image generated by our model, our model outperforms the baseline with a wide margin. On the VQA test set, in the case of BLEU measures, the improvement over the baseline grows from 24\% with unigram to 97\% with four-gram. It is evident that our model is capable to generate many more higher-order n-grams co-occurred in reference questions. This improvement is also consistent with ROUGE-L because it is based on the longest common subsequence between generated questions and reference questions. Our model performs better than the baseline also not just because it generates more exact higher order n-grams than reference questions. METEOR considers unigram alignment by allowing multiple matching modules to consider synonymy and alternating word forms. With this measure, our model is still 65\% higher than the baseline on the VQA test set. We also observe similar level of improvement over baseline on Visual7W dataset.

On both datasets, when the number of generated questions per image grows, the precison-oriented measures of our model are either similar or slightly declining because our model often generates meaningful questions that are not included in the ground-truth. The more questions we generate the more likely that the questions are not covered by manually constructed ones.

To measure the coverage of generated questions, we computed each reference question against all generated questions per image with all evaluation measures. As shown by Figure \ref{fig:PR_all}, all measures improves as the number of questions grows. Herein, both ROUGE-L and METEOR are way better than the baseline regardless of the number of generated questions on both datasets. When all six questions are generated, our model is 130\% better than the baseline across all measures. In particular, with METEOR, our model shows an improvement of 216\% and 179\% over the baseline on VQA and Visual7W respectively. When the number of manually constructed questions is small, our model provides even more types questions than manual ones, as shown with the examples in Figure \ref{Fig:Visual}. 

The distribution of question types generated by our model is more balanced than that of the ground-truth, while almost 55\% of questions in Visual7W and 89\% in VQA start with "\textit{what}", as illustrated by Figure \ref{Fig:question_types_dist}. Our model has also no tendency of generating too long or too short questions because the length distribution of the generated questions are very similar to that of the manually constructed datasets.

We also evaluate the effectiveness of the integration of bigram language model on both datasets. Herein, we compare two variants of our model, with and without the bigram model during decoding. As shown in Figure \ref{Fig:Ngrams}, regardless of precision or recall, decoding with the bigram model consistently outperforms the one without it. The inclusion of the bigram model effectively eliminates almost all repeated terms such as "the the" because the statistics collected by the bigram model favors grammatically well-formed sentences. This observation is also reflected in BLEU with higher-order ngrams by showing larger gaps.

\section{Conclusion}
In this paper, we propose the first model to automatically generate visually grounded questions with \textit{varying} types. Our model is capable of automatically selecting most likely question types and generating corresponding questions based on images and captions constructed by \textit{DenseCap}. 
Experiments on VQA and Visual7W dataset demonstrates that the proposed model is able to generate reasonable and grammatically well-formed questions with high diversity. 
For future work, we consider automatically generation of visual question-answer pairs, which will likely enhance training of VQA systems.

\section*{Acknowledgement}
This work is supported by National Key Technology R\&D Program of China: 2014BAK09B04, National Natural Science Foundation of China: U1636116, 11431006, Research Fund for International Young Scientists: 61650110510, Ministry of Education of Humanities and Social Science: 16YJC790123.

\scriptsize{
\bibliographystyle{named}
\bibliography{VL}

\begin{thebibliography}{}

\bibitem[\protect\citeauthoryear{Antol \bgroup \em et al.\egroup
  }{2015}]{Antol15}
Stanislaw Antol, Aishwarya Agrawal, Jiasen Lu, Margaret Mitchell, Dhruv Batra,
  C~Lawrence~Zitnick, and Devi Parikh.
\newblock Vqa: Visual question answering.
\newblock In {\em ICCV}, 2015.

\bibitem[\protect\citeauthoryear{Auer \bgroup \em et al.\egroup
  }{2007}]{Auer2007DBpediaAN}
S{\'o}ren Auer, Christian Bizer, Georgi Kobilarov, Jens Lehmann, Richard
  Cyganiak, and Zachary~G. Ives.
\newblock Dbpedia: A nucleus for a web of open data.
\newblock In {\em ISWC/ASWC}, 2007.

\bibitem[\protect\citeauthoryear{Banerjee and Lavie}{2005}]{banerjee2005meteor}
Satanjeev Banerjee and Alon Lavie.
\newblock Meteor: An automatic metric for mt evaluation with improved
  correlation with human judgments.
\newblock In {\em Acl workshop on intrinsic and extrinsic evaluation measures
  for machine translation and/or summarization}, 2005.

\bibitem[\protect\citeauthoryear{Barnard \bgroup \em et al.\egroup
  }{2003}]{Barnard03}
Kobus Barnard, Pinar Duygulu, David Forsyth, Nando~de Freitas, David~M Blei,
  and Michael~I Jordan.
\newblock Matching words and pictures.
\newblock {\em Journal of machine learning research}, 2003.

\bibitem[\protect\citeauthoryear{Bollacker \bgroup \em et al.\egroup
  }{2008}]{Bollacker2008FreebaseAC}
Kurt~D. Bollacker, Colin Evans, Praveen Paritosh, Tim Sturge, and Jamie Taylor.
\newblock Freebase: a collaboratively created graph database for structuring
  human knowledge.
\newblock In {\em SIGMOD}, 2008.

\bibitem[\protect\citeauthoryear{Bordes \bgroup \em et al.\egroup
  }{2014}]{bordes2014open}
Antoine Bordes, Jason Weston, and Nicolas Usunier.
\newblock Open question answering with weakly supervised embedding models.
\newblock In {\em ECML/PKDD}, pages 165--180, 2014.

\bibitem[\protect\citeauthoryear{Chen and Zitnick}{2014}]{Chen14}
Xinlei Chen and C~Lawrence Zitnick.
\newblock Learning a recurrent visual representation for image caption
  generation.
\newblock {\em arXiv preprint arXiv:1411.5654}, 2014.

\bibitem[\protect\citeauthoryear{Cho \bgroup \em et al.\egroup
  }{2014}]{cho2014learning}
Kyunghyun Cho, Bart Van~Merri{\"e}nboer, Caglar Gulcehre, Dzmitry Bahdanau,
  Fethi Bougares, Holger Schwenk, and Yoshua Bengio.
\newblock Learning phrase representations using rnn encoder-decoder for
  statistical machine translation.
\newblock {\em arXiv preprint arXiv:1406.1078}, 2014.

\bibitem[\protect\citeauthoryear{Chung \bgroup \em et al.\egroup
  }{2015}]{chung2015gated}
Junyoung Chung, Caglar G{\"u}l{\c{c}}ehre, Kyunghyun Cho, and Yoshua Bengio.
\newblock Gated feedback recurrent neural networks.
\newblock {\em CoRR, abs/1502.02367}, 2015.

\bibitem[\protect\citeauthoryear{Deng \bgroup \em et al.\egroup
  }{2009}]{deng2009imagenet}
Jia Deng, Wei Dong, Richard Socher, Li-Jia Li, Kai Li, and Li~Fei-Fei.
\newblock Imagenet: A large-scale hierarchical image database.
\newblock In {\em CVPR}, 2009.

\bibitem[\protect\citeauthoryear{Gao \bgroup \em et al.\egroup
  }{2015}]{gao2015you}
Haoyuan Gao, Junhua Mao, Jie Zhou, Zhiheng Huang, Lei Wang, and Wei Xu.
\newblock Are you talking to a machine? dataset and methods for multilingual
  image question.
\newblock In {\em Advances in Neural Information Processing Systems}, 2015.

\bibitem[\protect\citeauthoryear{Geman \bgroup \em et al.\egroup
  }{2015}]{Geman15}
Donald Geman, Stuart Geman, Neil Hallonquist, and Laurent Younes.
\newblock Visual turing test for computer vision systems.
\newblock {\em Proceedings of the National Academy of Sciences}, 2015.

\bibitem[\protect\citeauthoryear{Gupta and Mannem}{2012}]{Gupta12}
Ankush Gupta and Prashanth Mannem.
\newblock From image annotation to image description.
\newblock In {\em ICNIP}. Springer, 2012.

\bibitem[\protect\citeauthoryear{He \bgroup \em et al.\egroup
  }{2015}]{he2015delving}
Kaiming He, Xiangyu Zhang, Shaoqing Ren, and Jian Sun.
\newblock Delving deep into rectifiers: Surpassing human-level performance on
  imagenet classification.
\newblock In {\em ICCV}, 2015.

\bibitem[\protect\citeauthoryear{Hochreiter and
  Schmidhuber}{1997}]{hochreiter1997long}
Sepp Hochreiter and J{\"u}rgen Schmidhuber.
\newblock Long short-term memory.
\newblock {\em Neural computation}, 1997.

\bibitem[\protect\citeauthoryear{Hodosh \bgroup \em et al.\egroup
  }{2013}]{Hodosh13}
Micah Hodosh, Peter Young, and Julia Hockenmaier.
\newblock Framing image description as a ranking task: Data, models and
  evaluation metrics.
\newblock {\em Journal of Artificial Intelligence Research}, 2013.

\bibitem[\protect\citeauthoryear{Johnson \bgroup \em et al.\egroup
  }{2015}]{johnson2015densecap}
Justin Johnson, Andrej Karpathy, and Li~Fei-Fei.
\newblock Densecap: Fully convolutional localization networks for dense
  captioning.
\newblock {\em arXiv preprint arXiv:1511.07571}, 2015.

\bibitem[\protect\citeauthoryear{Karpathy and Fei-Fei}{2015}]{Karpathy15}
Andrej Karpathy and Li~Fei-Fei.
\newblock Deep visual-semantic alignments for generating image descriptions.
\newblock In {\em CVPR}, 2015.

\bibitem[\protect\citeauthoryear{Kingma and Ba}{2014}]{kingma2014adam}
Diederik Kingma and Jimmy Ba.
\newblock Adam: A method for stochastic optimization.
\newblock {\em arXiv preprint arXiv:1412.6980}, 2014.

\bibitem[\protect\citeauthoryear{Kneser and Ney}{1995}]{kneser1995improved}
Reinhard Kneser and Hermann Ney.
\newblock Improved backing-off for m-gram language modeling.
\newblock In {\em ICASSP-95}, 1995.

\bibitem[\protect\citeauthoryear{Koehn \bgroup \em et al.\egroup
  }{2003}]{Koehn2003}
Philipp Koehn, Franz~Josef Och, and Daniel Marcu.
\newblock Statistical phrase-based translation.
\newblock In {\em NAACL}, pages 48--54, 2003.

\bibitem[\protect\citeauthoryear{Kong \bgroup \em et al.\egroup
  }{2014}]{kong14}
Chen Kong, Dahua Lin, Mohit Bansal, Raquel Urtasun, and Sanja Fidler.
\newblock What are you talking about? text-to-image coreference.
\newblock In {\em CVPR}, 2014.

\bibitem[\protect\citeauthoryear{Krishna \bgroup \em et al.\egroup
  }{2016}]{krishnavisualgenome}
Ranjay Krishna, Yuke Zhu, Oliver Groth, Justin Johnson, Kenji Hata, Joshua
  Kravitz, Stephanie Chen, Yannis Kalantidis, Li-Jia Li, David~A Shamma,
  Michael Bernstein, and Li~Fei-Fei.
\newblock Visual genome: Connecting language and vision using crowdsourced
  dense image annotations.
\newblock 2016.

\bibitem[\protect\citeauthoryear{Kulkarni \bgroup \em et al.\egroup
  }{2011}]{kulkarni11}
Girish Kulkarni, Visruth Premraj, Sagnik Dhar, Siming Li, Yejin Choi,
  Alexander~C Berg, and Tamara~L Berg.
\newblock Baby talk: Understanding and generating image descriptions.
\newblock In {\em CVPR}, 2011.

\bibitem[\protect\citeauthoryear{Kuznetsova \bgroup \em et al.\egroup
  }{2012}]{kuznetsova12}
Polina Kuznetsova, Vicente Ordonez, Alexander~C Berg, Tamara~L Berg, and Yejin
  Choi.
\newblock Collective generation of natural image descriptions.
\newblock In {\em 50th Annual Meeting of the Association for Computational
  Linguistics}, 2012.

\bibitem[\protect\citeauthoryear{Kwiatkowski \bgroup \em et al.\egroup
  }{2013}]{Kwiatkowski2013}
Tom Kwiatkowski, Eunsol Choi, Yoav Artzi, and Luke Zettlemoyer.
\newblock Scaling semantic parsers with on-the-fly ontology matching.
\newblock In {\em EMNLP}, 2013.

\bibitem[\protect\citeauthoryear{Li \bgroup \em et al.\egroup }{2009}]{li09}
Li-Jia Li, Richard Socher, and Li~Fei-Fei.
\newblock Towards total scene understanding: Classification, annotation and
  segmentation in an automatic framework.
\newblock In {\em CVPR}, 2009.

\bibitem[\protect\citeauthoryear{Lin \bgroup \em et al.\egroup }{2014}]{lin14}
Tsung-Yi Lin, Michael Maire, Serge Belongie, James Hays, Pietro Perona, Deva
  Ramanan, Piotr Doll{\'a}r, and C~Lawrence Zitnick.
\newblock Microsoft coco: Common objects in context.
\newblock In {\em ECCV}, 2014.

\bibitem[\protect\citeauthoryear{Lin}{2004}]{lin2004rouge}
Chin-Yew Lin.
\newblock Rouge: A package for automatic evaluation of summaries.
\newblock In {\em Text summarization branches out: ACL-04 workshop}, 2004.

\bibitem[\protect\citeauthoryear{Malinowski and Fritz}{2014}]{malinowski14}
Mateusz Malinowski and Mario Fritz.
\newblock A multi-world approach to question answering about real-world scenes
  based on uncertain input.
\newblock In {\em Advances in Neural Information Processing Systems}, NIPS'14,
  2014.

\bibitem[\protect\citeauthoryear{Malinowski \bgroup \em et al.\egroup
  }{2015}]{malinowski15}
Mateusz Malinowski, Marcus Rohrbach, and Mario Fritz.
\newblock Ask your neurons: A neural-based approach to answering questions
  about images.
\newblock In {\em ICCV}, 2015.

\bibitem[\protect\citeauthoryear{Papineni \bgroup \em et al.\egroup
  }{2002}]{Papineni02bleu:a}
Kishore Papineni, Salim Roukos, Todd Ward, and Wei jing Zhu.
\newblock Bleu: a method for automatic evaluation of machine translation.
\newblock 2002.

\bibitem[\protect\citeauthoryear{Pennington \bgroup \em et al.\egroup
  }{2014}]{pennington2014glove}
Jeffrey Pennington, Richard Socher, and Christopher~D. Manning.
\newblock Glove: Global vectors for word representation.
\newblock In {\em Empirical Methods in Natural Language Processing (EMNLP)},
  2014.

\bibitem[\protect\citeauthoryear{Pirsiavash \bgroup \em et al.\egroup
  }{2014}]{pirsiavash14}
Hamed Pirsiavash, Carl Vondrick, and Antonio Torralba.
\newblock Inferring the why in images.
\newblock {\em arXiv preprint arXiv:1406.5472}, 2014.

\bibitem[\protect\citeauthoryear{Reddy \bgroup \em et al.\egroup
  }{2016}]{Reddy2016}
Siva Reddy, Oscar T{\'a}ckstr{\'0}m, Michael Collins, Tom Kwiatkowski, Dipanjan
  Das, Mark Steedman, and Mirella Lapata.
\newblock Transforming dependency structures to logical forms for semantic
  parsing.
\newblock In {\em TACL}, 2016.

\bibitem[\protect\citeauthoryear{Ren \bgroup \em et al.\egroup }{2015}]{ren15}
Mengye Ren, Ryan Kiros, and Richard Zemel.
\newblock Exploring models and data for image question answering.
\newblock In {\em Advances in Neural Information Processing Systems}, 2015.

\bibitem[\protect\citeauthoryear{Scott}{2008}]{scott2008kernel}
David~W Scott.
\newblock Kernel density estimators.
\newblock {\em Multivariate Density Estimation: Theory, Practice, and
  Visualization}, pages 125--193, 2008.

\bibitem[\protect\citeauthoryear{Simoncelli and
  Olshausen}{2001}]{MostafazadehMDZ16}
Eero~P Simoncelli and Bruno~A Olshausen.
\newblock Natural image statistics and neural representation.
\newblock {\em Annual review of neuroscience}, 2001.

\bibitem[\protect\citeauthoryear{Simonyan and Zisserman}{2014}]{Simonyan14c}
K.~Simonyan and A.~Zisserman.
\newblock Very deep convolutional networks for large-scale image recognition.
\newblock {\em CoRR}, abs/1409.1556, 2014.

\bibitem[\protect\citeauthoryear{Suchanek \bgroup \em et al.\egroup
  }{2007}]{Suchanek2007YagoAC}
Fabian~M. Suchanek, Gjergji Kasneci, and Gerhard Weikum.
\newblock Yago: a core of semantic knowledge.
\newblock In {\em WWW}, 2007.

\bibitem[\protect\citeauthoryear{Szegedy \bgroup \em et al.\egroup
  }{2015}]{szegedy2015going}
Christian Szegedy, Wei Liu, Yangqing Jia, Pierre Sermanet, Scott Reed, Dragomir
  Anguelov, Dumitru Erhan, Vincent Vanhoucke, and Andrew Rabinovich.
\newblock Going deeper with convolutions.
\newblock In {\em CVPR}, 2015.

\bibitem[\protect\citeauthoryear{Vinyals \bgroup \em et al.\egroup
  }{2015}]{vinyals15}
Oriol Vinyals, Alexander Toshev, Samy Bengio, and Dumitru Erhan.
\newblock Show and tell: A neural image caption generator.
\newblock In {\em CVPR}, 2015.

\bibitem[\protect\citeauthoryear{Weston \bgroup \em et al.\egroup
  }{2015}]{weston15}
Jason Weston, Antoine Bordes, Sumit Chopra, Alexander~M Rush, Bart van
  Merri{\"e}nboer, Armand Joulin, and Tomas Mikolov.
\newblock Towards ai-complete question answering: A set of prerequisite toy
  tasks.
\newblock {\em arXiv preprint arXiv:1502.05698}, 2015.

\bibitem[\protect\citeauthoryear{Xu \bgroup \em et al.\egroup
  }{2015}]{xu2015show}
Kelvin Xu, Jimmy Ba, Ryan Kiros, Kyunghyun Cho, Aaron Courville, Ruslan
  Salakhutdinov, Richard~S Zemel, and Yoshua Bengio.
\newblock Show, attend and tell: Neural image caption generation with visual
  attention.
\newblock {\em arXiv preprint arXiv:1502.03044}, 2015.

\bibitem[\protect\citeauthoryear{Yao and Durme}{2014}]{Yao2014InformationEO}
Xuchen Yao and Benjamin~Van Durme.
\newblock Information extraction over structured data: Question answering with
  freebase.
\newblock In {\em ACL}, 2014.

\bibitem[\protect\citeauthoryear{Yu \bgroup \em et al.\egroup }{2015}]{yu15}
Licheng Yu, Eunbyung Park, Alexander~C Berg, and Tamara~L Berg.
\newblock Visual madlibs: Fill in the blank image generation and question
  answering.
\newblock {\em arXiv preprint arXiv:1506.00278}, 2015.

\bibitem[\protect\citeauthoryear{Zhu \bgroup \em et al.\egroup
  }{2015}]{zhu2015visual7w}
Yuke Zhu, Oliver Groth, Michael Bernstein, and Li~Fei-Fei.
\newblock Visual7w: Grounded question answering in images.
\newblock {\em arXiv preprint arXiv:1511.03416}, 2015.

\bibitem[\protect\citeauthoryear{Zitnick \bgroup \em et al.\egroup
  }{2013}]{zitnick2013learning}
C~Lawrence Zitnick, Devi Parikh, and Lucy Vanderwende.
\newblock Learning the visual interpretation of sentences.
\newblock In {\em ICCV}, pages 1681--1688, 2013.

\end{thebibliography}
}

\end{document}